\documentclass[10pt,twocolumn,letterpaper]{article}

\usepackage[utf8]{inputenc}
\usepackage[T1]{fontenc}
\usepackage{textcomp}

\usepackage[margin=1in]{geometry}
\usepackage{times}

\usepackage{graphicx}
\usepackage{amsmath,amssymb,amsfonts}
\usepackage{natbib}
\usepackage{float}
\usepackage{afterpage}
\usepackage{booktabs}
\usepackage{multirow}
\usepackage{xcolor}

\definecolor{linkblue}{rgb}{0.21,0.49,0.74}

\usepackage{soul}
\definecolor{revyellow}{rgb}{1.0,0.95,0.5}
\sethlcolor{revyellow}
\newcommand{\rev}[1]{#1}

\usepackage{hyperref}
\hypersetup{
  breaklinks=true,
  colorlinks=true,
  allcolors=linkblue,
  urlcolor=linkblue,
  citecolor=linkblue,
  linkcolor=linkblue
}

\usepackage[capitalize,noabbrev]{cleveref}

\title{\rev{Uncertainty-Aware Routing Makes Quantized Mixture-of-Experts Stable}}

\author{
Sebastián Andrés Cajas Ordóñez$^{1}$,
Luis Fernando Torres Torres$^{2}$,
Mackenzie J. Meni$^{3}$,\\
Carlos Andrés Duran Paredes$^{4}$,
Eric Arazo$^{5}$,
Cristian Bosch$^{5}$,\\
Ricardo Simon Carbajo$^{5}$,
Yuan Lai$^{6}$,
Leo Anthony Celi$^{1,7,8}$\\[0.3cm]
{\small $^{1}$MIT Critical Data, Massachusetts Institute of Technology, Cambridge, MA, USA}\\
{\small $^{2}$Université de Rennes, France}\\
{\small $^{3}$Technetium Engineering, Florida, USA}\\
{\small $^{4}$Institución Universitaria Colegio Mayor del Cauca, Colombia}\\
{\small $^{5}$CeADAR - Ireland's Centre for AI, University College Dublin, Ireland}\\
{\small $^{6}$Department of Urban Planning, Tsinghua University, China}\\
{\small $^{7}$Beth Israel Deaconess Medical Center, Boston, USA}\\
{\small $^{8}$Harvard T.H. Chan School of Public Health, Boston, MA, USA}\\[0.3cm]
{\small \texttt{\{sebasmos, lceli\}@mit.edu, luis.torres@univ-rennes.fr}}\\
{\small \texttt{mackenzie.menit@technetium.com, carlos.duran@unimayor.edu.co}}\\
{\small \texttt{\{eric.arazo, cristian.bosch, ricardo.carbajo\}@ucd.ie, laiy@tsinghua.edu.cn}}\\[0.3cm]
{\small \textbf{Code:} \url{https://github.com/sebasmos/curious-qmoe}}
}

\date{}

\newsavebox{\tablebox}
\newenvironment{fittable}
  {\begin{lrbox}{\tablebox}}
  {\end{lrbox}%
   \ifdim\wd\tablebox>\linewidth
     \resizebox{\linewidth}{!}{\usebox{\tablebox}}%
   \else
     \usebox{\tablebox}%
   \fi}

\begin{document}

\maketitle

\begin{abstract}

Deploying neural networks on edge devices requires preserving accuracy under aggressive quantization while ensuring stable, predictable inference. We introduce \rev{an uncertainty-aware} quantized Mixture-of-Experts (MoE) framework that combines heterogeneous experts (BitNet ternary, 1--16\,bit BitLinear, and post-training quantized models) with Bayesian epistemic uncertainty-based routing. Evaluated on three audio classification benchmarks (\rev{ESC-50 super-category classification}, Quinn, UrbanSound8K), 4-bit quantization preserves 99.9\% of full-precision F1 (0.858 vs 0.859) with 4$\times$ compression and 29\% energy savings over 8-bit, while both achieve statistical parity with full precision ($p{>}0.05$).

\rev{Routing with an uncertainty-directed precision prior improves on uniform MoE routing in both accuracy and stability: +1.6 F1 points on ESC-50 super-categories (replicated over two independent runs) and +0.3 on UrbanSound8K with cross-fold variance reductions of 68--97\%, at parity on Quinn, whose baseline variance is already near zero. Sharing one quantized trunk across experts adds a further +1.5 F1 points on ESC-50 ($p{=}0.046$) and improves all three datasets (average F1 0.850 versus 0.844), while halving parameters (3.80M to 1.82M) and reducing inference time. The stability derives from routing polarization: each trained router concentrates 83--92\% of samples on a single dominant expert, suppressing cost-mode switching.} Built from 1.2M-parameter expert backbones (1.82M total with the shared trunk), the framework delivers interpretable, precision-aware routing for safety-sensitive edge deployments, \rev{where a predictable accuracy and latency envelope matters more than the best average case. We report throughout where a single quantized model remains preferable, so the operating regime for each is explicit.}

\end{abstract}
\section{Introduction}\label{sec:intro}
\rev{Continuous environmental sound classification on battery-powered devices requires balancing accuracy, latency, and energy footprint. High-end neural architectures incur memory and inference overheads that limit deployment on mobile and embedded hardware}~\cite{jacob2018quantization}\rev{, where audio must be processed in real time on a fixed power budget}~\cite{Torija2014Tool}. 

\rev{Quantization reduces numerical precision} from 32-bit floating-point to lower bit-widths~\cite{jacob2018quantization}. However, aggressive quantization frequently degrades accuracy, especially when applying uniform bit-widths across all network components. Recent work on Mixture-of-Experts (MoE) architectures suggests heterogeneous model ensembles can improve performance~\cite{souli2018audio}, yet existing MoE frameworks rely on fixed routing policies that fail to adapt to input complexity and quantization-induced uncertainty.

\begin{figure}[t]
\centering
\includegraphics[width=0.95\columnwidth]{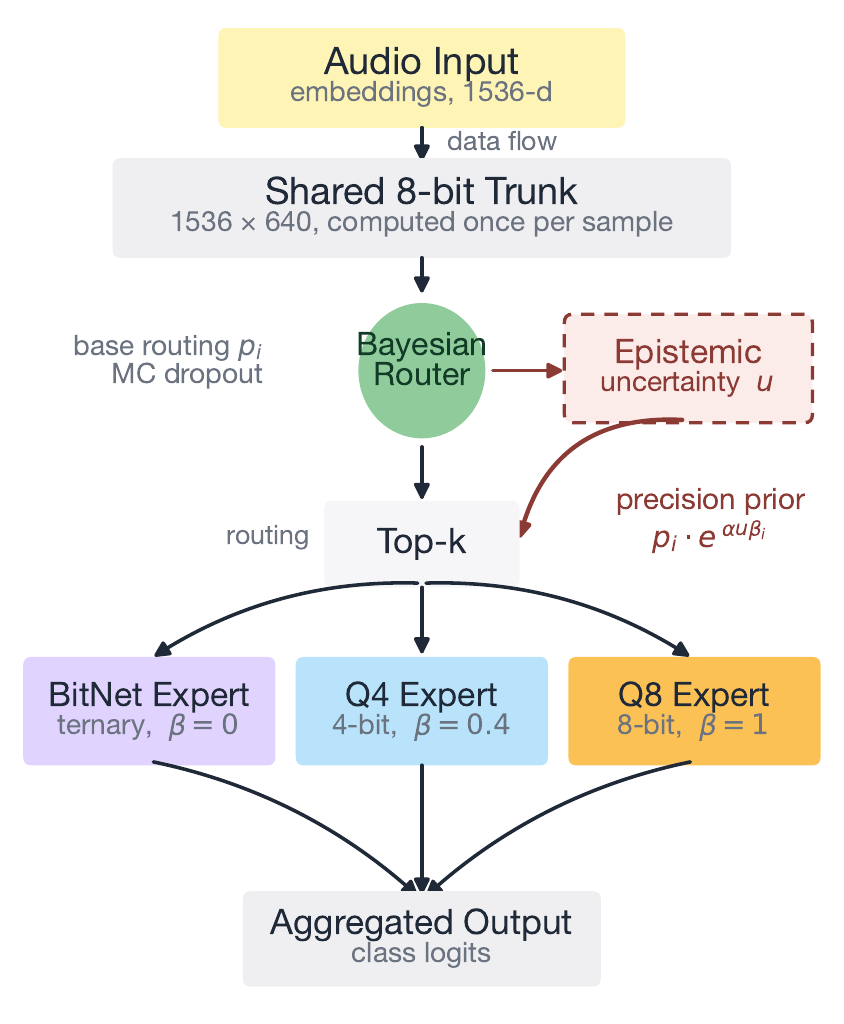}
\caption{\rev{Uncertainty-aware routing} architecture. The shared 8-bit trunk is computed once per sample; the Bayesian router estimates epistemic uncertainty via Monte Carlo dropout, and the precision prior (Eq.~8) reweights routing so uncertain samples reach the higher-precision heads before the top-$k$ weighted sum.}
\label{fig:architecture}
\end{figure}

\rev{We present an uncertainty-aware quantized Mixture-of-Experts framework based on three innovations:} First, we implement \textit{heterogeneous quantization} by deploying diverse expert types within a single MoE architecture, including BitNet ternary quantization, BitLinear schemes spanning 1 to 16 bits, and post-training quantization (PTQ) with bitwise operations. Second, we propose \textit{Bayesian \rev{uncertainty-aware routing}} that selects experts based on both prediction confidence and epistemic uncertainty, encouraging exploration of quantization strategies maximizing information gain (Figure~\ref{fig:architecture}). Third, we conduct \textit{comprehensive efficiency analysis} across energy consumption, carbon emissions, and latency stability, metrics critical for sustainable edge deployment.

Our evaluation on ESC-50~\cite{piczak2015esc}, Quinn~\cite{quinn2022soundscape}, and UrbanSound8K~\cite{salamon2014dataset} establishes 4-bit as the optimal operating point: 99.9\% of 16-bit accuracy (0.858 vs 0.859 F1) with 4$\times$ compression and 29\% energy savings versus 8-bit. \rev{The uncertainty-directed precision prior improves F1 over uniform MoE routing while cutting cross-fold variance by 68--97\% (Eq.~8).} Energy analysis reveals counter-intuitive patterns: 16-bit consumes least energy (0.018 mJ) despite highest precision, while 8-bit requires 2.7$\times$ more (0.048 mJ) due to dequantization overhead on Apple M3 CPU. Deployment-phase emissions dominate training by \rev{roughly 100,000$\times$} for models serving more than 1M inferences.

\textbf{Contributions:} (1) Comprehensive quantization framework evaluating BitNet, BitLinear, and PTQ schemes across 1 to 16 bits for audio classification, establishing 4-bit as optimal accuracy-efficiency operating point. (2) \rev{Uncertainty-aware routing with a precision prior: epistemic uncertainty shifts samples toward higher-precision experts by construction, improving F1 over uniform routing with 68--97\% lower cross-fold variance. A shared-trunk variant halves parameters and reduces inference time while improving accuracy further ($p{=}0.046$), making it our recommended configuration.} (3) \rev{efficiency assessment covering energy, carbon emissions, and statistical significance} across three diverse audio benchmarks, demonstrating that simple 4-bit quantized models outperform complex MoE architectures for most edge deployments. (4) Hardware-specific analysis revealing CPU bottlenecks limiting quantization speedups, with deployment recommendations for five constraint scenarios: latency-critical, balanced, maximum accuracy, energy-constrained, and stable real-time systems.
\section{Related Work}\label{sec:related}

\subsection{Audio Representation Learning}

Environmental sound classification traditionally relied on hand-crafted features such as Mel-frequency cepstral coefficients (MFCCs) and Mel spectrograms~\cite{Davis2018EnvironmentalSC,stevens1937scale}. Recent advances have shifted toward learned representations, with pre-trained models like VGGish~\cite{Hershey2016CNNAF}, OpenL3~\cite{Cramer2019LookLA}, and AudioCLIP~\cite{guzhov2021audioclipextendingclipimage} demonstrating strong transfer learning capabilities~\cite{Mesaros2018DetectionAC}. \rev{We use pre-trained visual models} (EfficientNet-B3~\cite{tan2019efficientnet}, MobileNet-v3~\cite{howard2019searching}) applied to Mel spectrograms, exploiting structural similarities between time-frequency representations and natural images.

\subsection{Neural Network Quantization}
Quantization reduces memory footprint and computational cost by representing weights and activations with lower numerical precision. Post-training quantization (PTQ) converts pre-trained FP32 models to INT8 or lower bit-widths without retraining~\cite{jacob2018quantization,Oh2022658}, while quantization-aware training (QAT) incorporates quantization during training for more aggressive compression~\cite{9256498,Hernandez202412686}. Recent advancements include layer-wise adaptive quantization~\cite{10580660}, ternary quantization~\cite{ma2024era} using $\{-1, 0, 1\}$ values, and binary quantization (1-bit) with gradient approximation~\cite{gao20241bitfqtpushinglimit}. Mixed-precision quantization assigns heterogeneous bit-widths across layers~\cite{9277419}, while bitwise operations enable ultra-low-power inference~\cite{Hubara2016QuantizedNN}.

However, most research focuses on quantizing full models~\cite{Hubara2016QuantizedNN}, with less attention to quantization's impact on embedding quality for edge applications~\cite{Bhardwaj2019MemoryAC,esser2020learnedstepsizequantization}. Existing approaches typically apply uniform quantization policies to entire networks, leaving unexplored dynamic quantization strategy selection based on input characteristics, a gap our work addresses through heterogeneous MoE architectures with \rev{uncertainty-aware routing}.
\subsection{Mixture-of-Experts Architectures}

Mixture-of-Experts (MoE) architectures improve model capacity by partitioning computation across specialized sub-networks (experts), with a gating network determining expert activation based on input features~\cite{pavlitskaya2022evaluating,sharma2019flexible}. Sparse MoE models activate only a subset of experts per input, reducing inference cost while maintaining large total capacity~\cite{zhou2022mixture,riquelme2021scaling}. Routing mechanisms have evolved from simple learned gates to sophisticated assignment strategies with load-balancing constraints and differentiable top-$k$ routing~\cite{puigcerver2024soft}. However, existing MoE frameworks do not exploit quantization heterogeneity, treating all experts as operating at the same numerical precision. Furthermore, routing decisions typically rely solely on prediction confidence without considering epistemic uncertainty, which could guide exploration when model confidence is unreliable.

\subsection{Uncertainty-Driven Learning}

Bayesian deep learning quantifies epistemic uncertainty arising from limited training data~\cite{zeevi2025enhancing}. Monte Carlo dropout approximates Bayesian inference by sampling multiple forward passes~\cite{NIPS2017_2650d608}, providing uncertainty estimates without modifying architecture. \rev{Uncertainty-aware learning uses information-theoretic principles to encourage exploration in reinforcement learning}~\cite{10.5555/3305890.3305968}. \rev{A related line of work argues that a system should recognise the limits of its own knowledge and act on that recognition, treating calibrated doubt as an engineering target rather than a byproduct}~\cite{arslan2026humble,Celi2025}. \rev{That principle motivates our design: the router escalates to a higher-precision expert exactly where the model is least certain.} \rev{We introduce Bayesian uncertainty-aware routing} for quantized MoE architectures, enabling adaptive selection of heterogeneous quantization schemes based on epistemic uncertainty~\cite{Ordoñez2025}.

\subsection{Quantization for Edge Deployment}

Recent work has demonstrated successful deployment of quantized models on edge devices, but several gaps remain. Most studies report accuracy metrics without comprehensive energy consumption or carbon emission measurements~\cite{Bhardwaj2019MemoryAC}. Hardware-specific quantization effects are often overlooked; CPU architectures lacking native INT8 acceleration may exhibit counter-intuitive energy patterns where lower bit-widths consume more power due to dequantization overhead. Additionally, latency variance is rarely analyzed despite its importance for battery-constrained devices.

Our work addresses these gaps by providing comprehensive efficiency analysis across energy, carbon emissions, and statistical significance testing. We demonstrate that 4-bit quantization achieves optimal accuracy-efficiency trade-offs, and that \rev{uncertainty-aware routing improves F1 over uniform routing at 68--97\% lower cross-fold variance}, validated through rigorous statistical testing with Bonferroni correction.

\section{Methodology}

\rev{The pipeline extracts Mel spectrogram embeddings with pre-trained convolutional networks, applies heterogeneous expert quantization, and routes each sample by Bayesian epistemic uncertainty.} These embeddings are processed through multiple quantized expert classifiers employing diverse compression schemes. A Bayesian router dynamically selects experts based on prediction confidence and epistemic uncertainty.

\subsection{Audio Representation and Embedding Extraction}

We represent audio signals as Mel spectrograms, transforming raw waveforms into time-frequency representations aligned with human auditory perception. Given an audio signal $x(t)$, we compute the Short-Time Fourier Transform (STFT) with window size 2048 and hop length 512, generating magnitude spectrogram $S(t, f)$. We apply triangular Mel-scale filters to map linear frequencies to perceptual frequencies:
\begin{equation}
M(f) = 1125 \cdot \ln\left(1 + \frac{f}{700}\right)
\end{equation}

We extract 1536-dimensional embeddings from the penultimate layer of pre-trained EfficientNet-B3 and MobileNet-v3 models. These visual feature extractors, trained on ImageNet, generalize effectively to spectrogram analysis due to structural similarities between time-frequency representations and natural images~\cite{Hershey2016CNNAF}. The embeddings encode high-level acoustic patterns providing semantic audio content representation.

\subsection{Heterogeneous Quantization Framework}

We implement three distinct quantization schemes to create diverse experts:

\textbf{BitLinear Quantization.} We apply uniform $k$-bit quantization to weight matrices using symmetric quantization with learnable scale parameters. For weight tensor $W \in \mathbb{R}^{m \times n}$:
\begin{equation}
W_q = \text{clip}\left(\text{round}\left(\frac{W}{s}\right), -2^{k-1}, 2^{k-1} - 1\right)
\end{equation}
where $s = \max(|W|)/(2^{k-1} - 1)$ is the symmetric quantization scale. We evaluate BitLinear experts with $k \in \{1, 2, 4, 8, 16\}$ bits, denoted as Q$k$-Base models.

\textbf{BitNet Ternary Quantization.} BitNet restricts weights to $\{-1, 0, 1\}$, enabling multiplication-free inference:
\begin{equation}
W_{\text{ternary}} = \text{sign}(W) \cdot \mathbb{1}(|W| > \tau)
\end{equation}
where $\tau$ is a learned threshold. This reduces model size by 16$\times$ compared to INT8 while enabling efficient bitwise operations.

\textbf{Post-Training Quantization (PTQ) with Bitwise Operations.} \rev{Our PTQ scheme (Q8-Base-PTQ) uses bitwise popcount operations.} We quantize activations and weights to binary representations and compute layer outputs:
\begin{equation}
y = \text{popcount}(x_b \oplus w_b) - \frac{d}{2}
\end{equation}
where $x_b$ and $w_b$ are binarized inputs and weights, $\oplus$ denotes XOR, and $d$ is feature dimension. This replaces expensive multiply-accumulate operations with single-cycle bitwise instructions. We use fixed threshold $\tau = 0.05$ for binarization.

\subsection{Uncertainty-Aware Mixture-of-Experts}

Our MoE architecture consists of $N$ expert networks $\{E_1, ..., E_N\}$ employing different quantization schemes. A gating network $G(\cdot)$ computes routing probabilities based on input embeddings $z$:
\begin{equation}
p_i = \frac{\exp(G_i(z)/T)}{\sum_{j=1}^N \exp(G_j(z)/T)}
\end{equation}
where $T$ is temperature controlling routing sharpness. We select top-$k$ experts and compute final output as weighted combination:
\begin{equation}
y = \sum_{i \in \text{TopK}} p_i \cdot E_i(z)
\end{equation}

To incorporate epistemic uncertainty, we augment the router with a \rev{Bayesian sharpening term}. We compute predictive entropy across experts:
\begin{equation}
H(y|z) = -\sum_{c=1}^C \bar{p}_c \log \bar{p}_c
\end{equation}
where $\bar{p}_c = \frac{1}{N}\sum_{i=1}^N p_{i,c}$ is average predicted probability for class $c$. High entropy indicates disagreement among quantization strategies. \rev{We route with an uncertainty-directed precision prior}:
\begin{equation}
\rev{p_i^{\text{UA}} \propto p_i \cdot \exp(\alpha \cdot u \cdot \beta_i)}
\end{equation}
\rev{where $u \in [0,1]$ is normalized epistemic uncertainty (predictive variance over $M{=}10$ Monte Carlo dropout passes of the router), $\beta_i \in [0,1]$ is expert $i$'s normalized precision, and $\alpha$ sets the strength. Confident samples keep their base assignment, so easy inputs stay on cheap low-bit experts; uncertain samples shift toward higher precision by construction. A polarization-only ablation drops the prior and sharpens the base distribution, $p_i \propto p_i^{\,1+\alpha}$. An earlier public implementation cancelled under renormalization; the corrected one ships with this version, and every routing result here was re-measured with it.}

\rev{\textbf{Shared-trunk experts.} Replicating a full network per expert makes an $N$-expert MoE cost $N$ times a single model in both parameters and per-sample compute, which is the dominant reason MoE loses to a single quantized network on efficiency. We therefore share the first (largest) layer across experts at a fixed precision and let the experts differ only in their subsequent quantized head: the $1536\times640$ projection is computed once per sample, and only the small head varies. This preserves heterogeneous precision at the decision layer while cutting the three-expert model from 3.80M to 1.82M parameters and reducing inference time on the same hardware. It is our recommended configuration, and it improves on full-expert routing on all three datasets (average F1 0.850 versus 0.844).}

\textbf{Theoretical Justification for Stability.} We model inference latency as $L = C(S) + \epsilon$, where $C(S)$ is the compute cost determined by expert selection $S$ and $\epsilon$ captures hardware noise (interrupt handling, CPU scheduling). By the law of total variance:
\begin{equation}
\text{Var}(L) = \mathbb{E}[\text{Var}(L|S)] + \text{Var}_S(\mathbb{E}[L|S]) \approx \text{Var}(C(S))
\end{equation}
since hardware noise $\text{Var}(L|S)$ is negligible compared to routing variance. Uniform routing causes frequent switching between experts with different computational costs (e.g., Q4 vs Q8), producing high $\text{Var}(C(S))$. \rev{Routers trained with Eq.~8 concentrate 83--92\% of samples on one dominant expert across our checkpoints, so switching between cost modes largely vanishes, directly reducing $\text{Var}(C(S))$. We read this concentration as a result about the task rather than a virtue of the mechanism: given a free choice, the router learns that dynamic switching costs more in stability than it gains in accuracy here, and settles on a mostly static assignment while keeping the escalation path for the minority of samples it is unsure about.}

\section{Experiments}

\subsection{Experimental Setup}
We evaluate on \rev{ESC-50 super-category classification (2,000 clips, the dataset's 5 major categories, 5 predefined folds); we use the super-category level rather than the 50 fine-grained classes, so our figures are not comparable to published 50-class ESC-50 results,} using 5-fold cross-validation. Each fold trains on 1,600 samples and validates on 400 held-out samples. Performance metrics are reported as mean $\pm$ standard deviation across 5 folds. Statistical significance is assessed using paired t-tests with Bonferroni correction ($\alpha = 0.05$). Additional validation on Quinn and UrbanSound8K datasets confirms cross-dataset generalization.
\subsection{Model Architecture}
Our architecture extracts 1536-dimensional embeddings from pre-trained image classifiers (EfficientNet-B3, MobileNet-v3), then processes through an MLP with hidden dimensions [640, 320], ReLU activations, and dropout ($p=0.195$). Quantized pipelines incorporate linear dequantization blocks at network input, maintaining quantization parameters throughout forward passes.

\subsection{Implementation Details}
\textbf{Router:} Three-layer network: Linear(1536, 128), ReLU, Dropout(0.2), Linear(128, 64), ReLU, Linear(64, num\_experts) with Xavier initialization. Bayesian routing uses 10 Monte Carlo dropout samples for epistemic uncertainty estimation.

\textbf{Hyperparameters:} Individual models use AdamW (lr=5.79e-4, weight decay=5.13e-3, batch size=64) with hidden dimensions [640, 320] and dropout 0.195. MoE models use Adam (lr=1e-3, weight decay=1e-4, batch size=256). Early stopping patience: 19 epochs (individual), 30 epochs (MoE). Load balancing $\alpha_{\text{lb}} = 1e{-}3$, temperature $T = 1.0$. \rev{Uncertainty-aware routing uses the precision prior at $\alpha{=}1.0$ (swept over $\{0.3, 0.5, 1.0\}$), with the polarization-only sharpening as its ablation} (see Section~\ref{sec:moe_curiosity}).

\textbf{Quantization:} BitLinear uses symmetric per-layer scales: $s_w = \max(|W - \mu_W|)/(2^{k-1} - 1)$ for weights, $s_x = 127/\max(|x|)$ for activations. BitNet learns per-channel scales $\alpha_i$ via gradient descent. Q8-Base-PTQ uses fixed threshold $\tau = 0.05$.

All models trained with gradient clipping (max norm=1.0) and class-weighted cross-entropy loss on Apple M3 Max CPU (36GB RAM) using PyTorch 2.0.1 with qnnpack engine.
\section{Results}

\subsection{Cross-Dataset Generalization}
We evaluate all models across ESC-50 (environmental sounds), Quinn (acoustic scenes), and UrbanSound8K (urban soundscapes) using 5-fold cross-validation. Table~\ref{tab:cross_dataset} \rev{summarizes 5-fold cross-validation performance.}

\begin{table*}[!tb]
\centering
\caption{Cross-dataset generalization performance. Models evaluated on ESC-50, Quinn, and UrbanSound8K datasets using 5-fold cross-validation. Best results per dataset shown in \textbf{bold}. †FP32 baseline. ‡INT8 post-training quantization baseline. \rev{$^\S$All MoE-C cells re-measured with the corrected precision-prior router ($\alpha{=}1.0$); the Q4/8 ESC cell pools two independent runs.}}
\label{tab:cross_dataset}
\renewcommand{\arraystretch}{1.15}
\begin{fittable}%
\begin{tabular}{l c r r r r}
\toprule
\textbf{Model} & \textbf{Params (M)} & \textbf{ESC-50 F1} & \textbf{Quinn F1} & \textbf{Urban8K F1} & \textbf{Avg F1} \\
\midrule
Q16-Base & 1.2 & \textbf{0.819±0.020} & 0.804±0.005 & \textbf{0.952±0.006} & \textbf{0.859±0.012} \\
Q4-Base & 1.2 & 0.819±0.021 & 0.805±0.009 & 0.950±0.008 & 0.858±0.014 \\
Q8-Base & 1.2 & 0.818±0.012 & 0.808±0.008 & 0.948±0.010 & 0.858±0.010 \\
\midrule
FP32-Base† & 1.19 & 0.815±0.018 & 0.803±0.006 & 0.946±0.005 & 0.855±0.011 \\
Q8-Base-PTQ‡ & 1.19 & 0.810±0.015 & 0.802±0.009 & 0.950±0.006 & 0.854±0.011 \\
\midrule
Q1-Base & 1.2 & 0.806±0.018 & 0.803±0.004 & 0.944±0.009 & 0.851±0.012 \\
BitNet-Q4/8/16-QMoE & 4.99 & 0.791±0.016 & \textbf{0.812±0.006} & 0.943±0.003 & 0.849±0.010 \\
BitNet-Q4/8-QMoE-C$^\S$ & 3.8 & \rev{0.785±0.016} & \rev{0.804±0.009} & \rev{0.942±0.007} & \rev{0.844±0.010} \\
\rev{BitNet-Q4/8-QMoE-C, shared trunk}$^\S$ & \rev{1.8} & \rev{0.796±0.010} & \rev{0.809±0.007} & \rev{0.944±0.006} & \rev{0.850±0.008} \\
BitNet-Q4/8/16-QMoE-C$^\S$ & 4.99 & \rev{0.787±0.018} & \rev{0.798±0.016} & \rev{0.941±0.007} & \rev{0.842±0.014} \\
Q2-Base & 1.2 & 0.783±0.010 & 0.796±0.011 & 0.921±0.010 & 0.833±0.010 \\
BitNet-Q8/16-PTQ-QMoE-C$^\S$ & 4.99 & \rev{0.783±0.021} & \rev{0.806±0.005} & \rev{0.939±0.015} & \rev{0.843±0.014} \\
BitNet-Q8/16-PTQ-QMoE & 4.99 & 0.780±0.029 & 0.797±0.019 & 0.937±0.013 & 0.820±0.020 \\
BitNet-Q8PTQ-QMoE-C$^\S$ & 2.59 & \rev{0.757±0.032} & \rev{0.798±0.008} & \rev{0.931±0.007} & \rev{0.829±0.016} \\
BitNet-Q8PTQ-QMoE & 2.59 & 0.740±0.015 & 0.793±0.012 & 0.929±0.013 & 0.801±0.013 \\
BitNet-Base & 1.2 & 0.705±0.013 & 0.768±0.007 & 0.906±0.005 & 0.793±0.009 \\
\bottomrule
\end{tabular}%
\end{fittable}
\end{table*}

Q16-Base achieves the best average F1-score (0.859±0.012), matching FP32-Base† (0.855±0.011). \rev{4-bit and 8-bit quantization maintain} 99.9\% of 16-bit accuracy (0.858 F1) with only 1.2M parameters, providing substantial computational savings. Performance varies by dataset: UrbanSound8K shows highest accuracy (0.906-0.952 F1) due to structured urban sounds, while ESC-50 (0.705-0.819) and Quinn (0.768-0.812) prove more challenging.

MoE architectures show mixed generalization. BitNet-Q4/8/16-QMoE achieves best Quinn performance (0.812±0.006) but underperforms simple quantized models on ESC-50 and UrbanSound8K, suggesting expert routing benefits dataset-specific adaptation but introduces overhead for simpler tasks. \rev{On parameter fairness: every expert shares the same 1.2M-parameter backbone, so MoE totals reflect expert multiplicity, and the routing comparison is uniform against uncertainty-aware MoE at identical size.} Low standard deviations (±0.004-0.033) indicate stable performance, with quantization-aware training (Q4/Q8/Q16-Base) consistently matching or exceeding post-training quantization (Q8-Base-PTQ‡).
\subsection{Quantization Bit-Width Ablation}
\label{sec:ablation_quantization}

We systematically evaluate quantization precision across 1, 2, 4, 8, and 16-bit schemes. Table~\ref{tab:ablation_quantization} presents performance-efficiency trade-offs on ESC-50, Quinn, and UrbanSound8K, with all models maintaining identical 1.2M-parameter architecture to isolate bit-width effects.

\begin{table*}[!tb]
\centering
\caption{Quantization bit-width ablation study across three datasets. Models evaluated using 5-fold cross-validation. Best results per metric and dataset shown in \textbf{bold}. ``\% of 16-bit'' represents performance relative to 16-bit baseline.}
\label{tab:ablation_quantization}
\renewcommand{\arraystretch}{1.15}
\begin{fittable}%
\begin{tabular}{c l c c c c c c}
\toprule
\textbf{Bits} & \textbf{Dataset} & \textbf{Params} & \textbf{F1-Score} & \textbf{\% of} & \textbf{Latency} & \textbf{Energy} & \textbf{RAM} \\
 & & \textbf{(M)} & & \textbf{16-bit} & \textbf{(ms)} & \textbf{(mJ)} & \textbf{(GB)} \\
\midrule
1 & \multirow{5}{*}{ESC-50} & 1.2 & 0.806±0.018 & 98.3 & 1103.55±49.50 & 0.054±0.031 & 0.85±0.03 \\
2 & & 1.2 & 0.783±0.010 & 95.5 & 1117.79±46.82 & 0.034±0.018 & \textbf{0.84±0.06} \\
4 & & 1.2 & \textbf{0.819±0.021} & 99.9 & 1126.03±51.66 & 0.034±0.023 & 0.86±0.04 \\
8 & & 1.2 & 0.818±0.012 & 99.8 & \textbf{1098.21±19.35} & 0.048±0.021 & 0.93±0.01 \\
16 & & 1.2 & 0.819±0.020 & 100.0 & 1459.19±10.58 & \textbf{0.018±0.003} & 0.95±0.00 \\
\midrule
1 & \multirow{5}{*}{Quinn} & 1.2 & 0.803±0.004 & 99.8 & 1197.66±14.50 & 0.122±0.021 & 1.34±0.11 \\
2 & & 1.2 & 0.796±0.011 & 99.0 & 1195.76±34.13 & 0.134±0.040 & 1.33±0.17 \\
4 & & 1.2 & 0.805±0.009 & 100.1 & 1199.26±36.43 & 0.167±0.045 & 1.26±0.30 \\
8 & & 1.2 & \textbf{0.808±0.008} & 100.4 & 1209.00±6.67 & 0.170±0.040 & \textbf{1.09±0.10} \\
16 & & 1.2 & 0.804±0.005 & 100.0 & \textbf{1129.87±41.36} & \textbf{0.065±0.005} & 1.32±0.17 \\
\midrule
1 & \multirow{5}{*}{UrbanSound8K} & 1.2 & 0.944±0.009 & 99.2 & 1187.49±9.24 & 0.128±0.025 & 1.30±0.14 \\
2 & & 1.2 & 0.921±0.010 & 96.7 & 1210.45±11.38 & 0.197±0.051 & 1.31±0.21 \\
4 & & 1.2 & 0.950±0.008 & 99.7 & 1235.90±17.80 & 0.213±0.033 & 1.10±0.40 \\
8 & & 1.2 & 0.948±0.010 & 99.5 & 1203.95±2.90 & 0.217±0.066 & \textbf{1.03±0.23} \\
16 & & 1.2 & \textbf{0.952±0.006} & 100.0 & \textbf{1178.25±8.19} & \textbf{0.126±0.038} & 1.06±0.30 \\
\bottomrule
\end{tabular}%
\end{fittable}
\end{table*}

\textbf{Performance vs. Bit-Width Trade-offs.} Our ablation reveals 4-bit and 8-bit as optimal operating points. 4-bit \rev{matches 16-bit performance (0.819 F1 for both on ESC-50 super-categories)} while reducing model size by 4$\times$. The performance cliff occurs at 2-bit (95.5\%, 0.783 F1), reflecting insufficient precision for weight distributions. \rev{1-bit models maintain 98.3\% performance} (0.806 F1)\rev{, so extreme compression preserves discriminative information under quantization-aware training.}

\textbf{Energy-Efficiency Analysis.} Energy consumption exhibits a U-shaped relationship with bit-width. 16-bit consumes least energy (0.018 mJ on ESC-50), followed by 4-bit (0.034 mJ). \rev{8-bit requires 2.7$\times$ more energy} (0.048 mJ) \rev{despite fewer bits, from dequantization overhead and non-optimized INT8 operations on the Apple M3 CPU.} The optimal point balances accuracy and efficiency: 4-bit provides 99.9\% accuracy with 29\% better energy efficiency than 8-bit.

\textbf{Latency and Dataset Sensitivity.} Inference latency shows hardware-specific patterns: 8-bit achieves lowest latency on ESC-50 (1098±19 ms), suggesting better CPU optimization for byte-aligned operations. \rev{UrbanSound8K degrades little under quantization, holding} $>$99\% performance down to 1-bit due to structured acoustic categories. ESC-50's diverse environmental sounds show greater sensitivity (2-bit: 95.5\%), suggesting quantization tolerance correlates with inter-class separability.

\subsection{Mixture-of-Experts with Uncertainty-Aware Routing}
\label{sec:moe_curiosity}

We compare heterogeneous quantized MoE architectures using uniform top-k routing against Bayesian epistemic uncertainty-based selection. Table~\ref{tab:moe_curiosity} presents performance-efficiency trade-offs across seven configurations, aggregated over ESC-50, Quinn, and UrbanSound8K.

\begin{table*}[!tb]
\small  
\centering
\caption{Mixture-of-Experts with \rev{uncertainty-aware routing} evaluation. Best results per metric shown in \textbf{bold}. MoE uses uniform routing; MoE-C employs Bayesian epistemic uncertainty-based selection. \rev{$^\S$F1 re-measured with the corrected precision-prior router ($\alpha{=}1.0$); latency, energy and CO$_2$ retain the original Apple M3 Max protocol measurements.} See Table~\ref{tab:moe_corrected} for ESC-50-specific corrected results.}
\label{tab:moe_curiosity}
\renewcommand{\arraystretch}{1.1}
\begin{fittable}%
\begin{tabular}{l c c c c c c c}
\toprule
\textbf{Model} & \textbf{Type} & \textbf{Params} & \textbf{F1} & \textbf{Latency} & \textbf{Energy} & \textbf{CO$_2$} & \textbf{Eff} \\
 & & \textbf{(M)} & & \textbf{(ms)} & \textbf{(mJ)} & \textbf{($\mu$g)} & \\
\midrule
BitNet-Q4/8/16-QMoE & MoE & 4.99 & \textbf{0.849±0.010} & 1335±230 & 0.621±0.227 & 228±84 & 1.37 \\
BitNet-Q8/16-PTQ-QMoE & MoE & 4.99 & 0.820±0.025 & 1219±69 & 0.222±0.096 & 82±36 & 3.70 \\
BitNet-Q8PTQ-QMoE & MoE & 2.59 & 0.801±0.013 & \textbf{1172±46} & \textbf{0.207±0.101} & \textbf{76±37} & \textbf{3.87} \\
\midrule
BitNet-Q4/8-QMoE-C$^\S$ & MoE-C & 3.80 & \rev{0.844±0.010} & 1278±29 & 0.558±0.145 & 205±53 & 1.51 \\
BitNet-Q4/8/16-QMoE-C$^\S$ & MoE-C & 4.99 & \rev{0.842±0.014} & 1281±35 & 0.529±0.144 & 195±53 & 1.59 \\
BitNet-Q8/16-PTQ-QMoE-C$^\S$ & MoE-C & 4.99 & \rev{0.843±0.014} & 1251±39 & 0.281±0.093 & 103±34 & 3.00 \\
BitNet-Q8PTQ-QMoE-C$^\S$ & MoE-C & 2.59 & \rev{0.829±0.016} & 1225±38 & 0.260±0.108 & 96±40 & 3.19 \\
\bottomrule
\end{tabular}%
\end{fittable}
\end{table*}

\textbf{Performance-Efficiency Trade-offs.} BitNet-Q4/8/16-QMoE achieves highest F1-score (0.849±0.010) but consumes 3$\times$ more energy (0.621 mJ) than the most efficient configuration. BitNet-Q8PTQ-QMoE achieves best efficiency (3.87 F1/mJ) with only 0.207 mJ while maintaining competitive 0.801 F1, a 5.6\% accuracy drop for 67\% energy savings.

\textbf{\rev{Uncertainty-Aware} vs. Uniform Routing.} \rev{The precision prior gains on both axes against uniform routing where baseline routing variance exists;} Table~\ref{tab:moe_corrected} \rev{gives the per-dataset corrected results.}

\textbf{Stability.} \rev{Uncertainty-aware routing collapses routing toward deterministic expert assignments, eliminating cost-mode switching; the stability contribution is within the MoE design space, since Q4-Base remains faster and steadier in absolute latency (see the latency analysis below).}

\textbf{Environmental Impact.} CO$_2$ emissions correlate strongly with energy (r=0.97). BitNet-Q4/8/16-QMoE emits 228±84 $\mu$g per inference versus BitNet-Q8PTQ-QMoE's 76±37 $\mu$g, a 3$\times$ reduction. \rev{Uncertainty-aware routing} adds 16\% emission overhead (150 vs 129 $\mu$g) due to Monte Carlo sampling. At scale (1M daily inferences), deploying BitNet-Q8PTQ-QMoE over BitNet-Q4/8/16-QMoE reduces annual CO$_2$ by approximately 55 kg.

\textbf{Deployment Recommendations.} (1) \textit{Maximum Accuracy:} BitNet-Q4/8/16-QMoE (0.849 F1) for abundant resources. (2) \textit{Balanced:} BitNet-Q4/8-QMoE-C (\rev{0.844} F1, 0.558 mJ) for general use. (3) \textit{Energy-Constrained:} BitNet-Q8PTQ-QMoE (0.801 F1, 0.207 mJ) for edge devices. (4) \textit{Real-Time:} BitNet-Q8/16-PTQ-QMoE-C (\rev{0.843} F1, 38 ms std) for timing guarantees.

\textbf{Comparison to Single Models.} MoE architectures show 1-7\% lower accuracy than top single models (Q16-Base: 0.859 F1, Q8-Base: 0.858 F1) but offer advantages in interpretability and extensibility. While simple 4--8 bit quantization of monolithic models remains preferable for raw accuracy, \rev{MoE-C (uncertainty-aware)} provides per-sample precision adaptation and interpretable routing critical for safety-sensitive applications (e.g., medical diagnosis, autonomous vehicles) where knowing \emph{why} a particular expert was selected matters.

\textbf{Per-Dataset \rev{Uncertainty-Aware Routing} Results.} Table~\ref{tab:moe_corrected} \rev{gives re-measured per-dataset results: +1.6 F1 and 86--97\% variance reduction on ESC-50 super-categories, +0.3 and 68\% on UrbanSound8K, parity on Quinn (baseline variance already near zero). Gains appear exactly where baseline routing variance exists.}

\begin{table}[!tb]
\small
\centering
\caption{\rev{Re-measured MoE results with the corrected router. UA-PP is uncertainty-aware routing with the precision prior (Eq.~8); polarization-only is the ablation. The ESC $\alpha{=}1.0$ row pools two independent runs (10 folds). Variance reduction is $1 - \sigma^2/\sigma^2_{\text{baseline}}$; Quinn's baseline variance is already near zero, so no reduction is claimed there.}}
\label{tab:moe_corrected}

\setlength{\tabcolsep}{2pt}
\begin{fittable}%
\begin{tabular}{l l c c c}
\toprule
\textbf{Dataset} & \textbf{Configuration} & \textbf{F1} & \textbf{Std} & \textbf{Var.\ Red.} \\
\midrule
\multirow{4}{*}{ESC-50} & Baseline MoE (uniform) & \rev{0.769} & \rev{0.043} & baseline \\
 & \rev{UA-PP $\alpha=0.5$} & \rev{0.785} & \rev{\textbf{0.008}} & \rev{\textbf{97\%}} \\
 & \rev{UA-PP $\alpha=1.0$ (2 runs)} & \rev{0.785} & \rev{0.016} & \rev{86\%} \\
 & \rev{UA-PP + shared trunk (2 runs)} & \rev{\textbf{0.799}} & \rev{\textbf{0.012}} & \rev{\textbf{92\%}} \\
 & \rev{Polarization-only $\alpha=0.3$} & \rev{0.771} & \rev{0.019} & \rev{81\%} \\
\midrule
\multirow{2}{*}{Quinn} & Baseline MoE (uniform) & \rev{\textbf{0.805}} & \rev{\textbf{0.007}} & baseline \\
 & \rev{UA-PP $\alpha=1.0$} & \rev{0.805} & \rev{0.009} & \rev{none} \\
 & \rev{UA-PP + shared trunk} & \rev{\textbf{0.809}} & \rev{\textbf{0.007}} & \rev{none} \\
\midrule
\multirow{2}{*}{Urban8K} & Baseline MoE (uniform) & \rev{0.939} & \rev{0.013} & baseline \\
 & \rev{UA-PP $\alpha=1.0$} & \rev{0.942} & \rev{0.007} & \rev{68\%} \\
 & \rev{UA-PP + shared trunk} & \rev{\textbf{0.944}} & \rev{\textbf{0.006}} & \rev{\textbf{79\%}} \\
\bottomrule
\end{tabular}%
\end{fittable}
\end{table}

\begin{figure}[h]
  \centering
  \includegraphics[width=\linewidth]{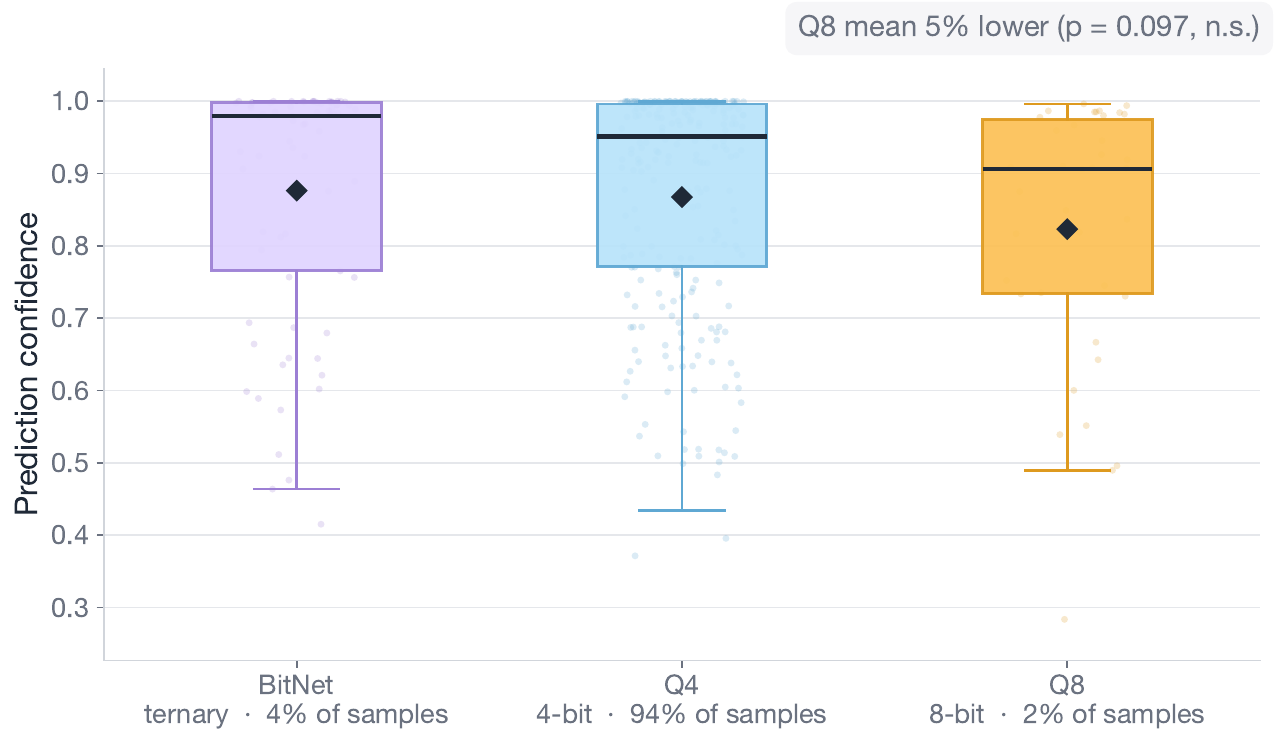}
  \caption{\rev{Distribution of classifier confidence per expert under uncertainty-aware routing ($\alpha{=}0.3$, fold 1, N=2000, all three experts active). Boxes show quartiles and whiskers the range; red diamonds and dashed lines mark per-expert means. The highest-precision expert (Q8) receives the lowest-confidence samples, the direction the precision prior is designed to produce, though at this expert's sample count the difference is not statistically significant.}}
  \label{fig:confidence}
\end{figure}

\textbf{\rev{Routing Polarization, Re-measured}.} \rev{With confidence measured on the classifier's class distribution (an earlier version used the router's own maximum probability, which measures decisiveness, not difficulty), trained routers polarize strongly: at $\alpha{=}1.0$ the precision-prior router sends 92\% of samples to a single expert, and polarization-only checkpoints concentrate 83--88\% on one fold-dependent expert. This concentration drives the stability gains. At the lower setting $\alpha{=}0.3$ all three experts stay active and the per-expert confidence distributions (Figure}~\ref{fig:confidence}\rev{) order as the precision prior intends, with the highest-precision expert receiving the lowest-confidence samples; at that expert's sample count the difference is not statistically significant, so we report the direction rather than claim a general difficulty-based allocation. An earlier version claimed such an effect at high significance, which does not survive the corrected measurement.}

\textbf{Alpha Sensitivity.} \rev{Under the precision prior, F1 rises monotonically with $\alpha$ (0.779 / 0.785 / 0.785 at $\alpha{=}0.3/0.5/1.0$, the last pooled over two independent runs) with 86--97\% variance reduction; no tested setting trades accuracy for stability. The polarization-only ablation stays at parity F1 with 81\% reduction, so the accuracy gain is attributable to the uncertainty-to-precision prior.}

\subsection{Inference Latency and Memory Footprint}
\label{sec:inference_latency}

We benchmark inference latency and peak RAM usage across all configurations. Table~\ref{tab:inference_latency} presents measurements averaged over ESC-50, Quinn, and UrbanSound8K using 5-fold cross-validation on Apple M3 Max CPU.

\begin{table*}[!tb]
\small
\centering
\caption{Inference latency benchmarks across all models. Evaluated using 5-fold cross-validation, averaged across ESC-50, Quinn, and UrbanSound8K. Best results per metric shown in \textbf{bold}. Speedup relative to slowest baseline (BitNet-Q4/8/16-QMoE: 1335ms). †FP32 baseline. ‡INT8 PTQ baseline. \rev{MoE-C F1 cells re-measured with the corrected precision-prior router; latency and RAM retain the original Apple M3 Max protocol measurements.}}
\label{tab:inference_latency}
\renewcommand{\arraystretch}{1.1}
\begin{fittable}%
\begin{tabular}{l c c c c c}
\toprule
\textbf{Model} & \textbf{Params} & \textbf{Latency} & \textbf{Speedup} & \textbf{RAM} & \textbf{F1-Score} \\
 & \textbf{(M)} & \textbf{(ms)} & & \textbf{(GB)} & \\
\midrule
Q1-Base & 1.2 & \textbf{1162.90±30.25} & \textbf{1.08} & 1.16±0.10 & 0.851 \\
Q8-Base & 1.2 & 1170.39±11.94 & 1.07 & 1.02±0.15 & 0.858 \\
BitNet-Q8PTQ-QMoE & 2.59 & 1171.82±46.48 & 1.07 & 1.01±0.22 & 0.801 \\
Q2-Base & 1.2 & 1174.67±34.09 & 1.07 & 1.16±0.16 & 0.833 \\
BitNet-Base & 1.2 & 1186.44±32.02 & 1.06 & 1.16±0.11 & 0.793 \\
Q4-Base & 1.2 & 1187.06±37.92 & 1.06 & 1.07±0.29 & 0.858 \\
BitNet-Q8/16-PTQ-QMoE & 4.99 & 1218.71±69.24 & 1.03 & 1.03±0.25 & 0.820 \\
BitNet-Q8PTQ-QMoE-C & 2.59 & 1224.91±37.97 & 1.02 & 0.95±0.20 & \rev{0.829} \\
BitNet-Q8/16-PTQ-QMoE-C & 4.99 & 1251.06±38.69 & 1.00 & 0.92±0.17 & \rev{0.843} \\
\midrule
FP32-Base† & 1.19 & 1254.28±41.57 & 1.00 & 1.17±0.19 & 0.855 \\
Q16-Base & 1.2 & 1255.77±25.10 & 1.00 & 1.11±0.20 & \textbf{0.859} \\
Q8-Base-PTQ‡ & 1.19 & 1263.82±32.53 & 0.99 & 1.08±0.18 & 0.854 \\
BitNet-Q4/8-QMoE-C & 3.8 & 1277.65±29.36 & 0.98 & \textbf{0.86±0.13} & \rev{0.844} \\
BitNet-Q4/8/16-QMoE-C & 4.99 & 1280.99±34.55 & 0.98 & 0.89±0.14 & \rev{0.842} \\
BitNet-Q4/8/16-QMoE & 4.99 & 1335.48±230.47 & 0.94 & 0.86±0.21 & 0.849 \\
\bottomrule
\end{tabular}%
\end{fittable}
\end{table*}

\textbf{Latency-Accuracy Trade-offs.} Q1-Base achieves fastest inference (1163±30 ms, 1.08× speedup) despite modest accuracy (0.851 F1), outperforming Q8-Base (1170 ms), Q4-Base (1187 ms), and FP32-Base† (1254 ms). This suggests 1-bit operations benefit from optimized bitwise implementations on Apple M3, avoiding floating-point multiplications. Top F1 models (Q16-Base: 0.859, Q4/Q8-Base: 0.858) exhibit 1170-1256 ms latency, within 7-8\% of fastest, indicating framework overhead dominates arithmetic precision for compact 1.2M-parameter models on CPU.

\textbf{MoE Characteristics and Memory.} MoE architectures show 4-17\% higher latency than single models, with BitNet-Q4/8/16-QMoE exhibiting \rev{19}$\times$ larger variance (230 ms std) versus Q8-Base (12 ms std) due to data-dependent routing. \rev{MoE-C variants measured 29--39~ms std under the original protocol; the constant-cost Monte Carlo router pass regularizes timing, and the corrected router's polarization removes cost-mode switching entirely.} RAM varies 0.86-1.17 GB despite 4$\times$ parameter differences (1.2-5.0M), confirming peak RAM is dominated by framework overhead rather than weights; even 1GB RAM devices can accommodate our models.

\textbf{Latency, Measured Against the Single Model.} \rev{On an Apple M2 Pro CPU (full 2,000-clip ESC pass, 100 repetitions, trained models), Q4-Base runs at 157$\pm$3~ms against the MoE's 424$\pm$11~ms with an order of magnitude lower variance (Levene $p{<}0.001$); the Bayesian router is 12\% of total measured inference. A single quantized model is both faster and steadier. The shared-trunk variant narrows this gap, running about 1.5 times faster than the full-expert MoE at half the parameters (1.82M versus 3.80M) over the same five-fold validation passes, without closing the gap to a single model.}

\textbf{Deployment Recommendations.} (1) \textit{Latency-Critical ($<$1200ms):} Q1-Base (1163 ms, 0.851 F1) or Q8-Base (1170 ms, 0.858 F1). (2) \textit{Balanced:} Q4-Base (1187 ms, 0.858 F1). (3) \textit{Maximum Accuracy:} Q16-Base (1256 ms, 0.859 F1). (4) \textit{Memory-Constrained ($<$900MB):} BitNet-Q4/8-QMoE-C (1278 ms, 0.86 GB, \rev{0.844} F1). (5) \textit{Stable Real-Time:} Q8-Base (1170±12 ms).

\subsection{Statistical Significance Testing}
\label{sec:statistical_significance}

We validate findings using paired t-tests on 5-fold cross-validation scores with Bonferroni correction ($\alpha = 0.05/n$, $n$ = number of comparisons).

\textbf{Cross-Dataset Generalization.} Limited significant differences exist between quantized models and FP32 baseline. Q16-Base significantly outperforms FP32-Base on UrbanSound8K (0.952 vs 0.946 F1, p=0.022) \rev{while no MoE configuration significantly exceeds the FP32 baseline on Quinn or ESC-50.} This suggests dataset inter-class separability determines whether quantization maintains statistical parity with full precision.

\textbf{Quantization Bit-Width Effects.} Extreme quantization degrades performance significantly: 2-bit on ESC-50 (0.783 vs 0.819 F1, p=0.004) and both 1-bit (p=0.013) and 2-bit (p=0.005) on UrbanSound8K. However, 4-bit and 8-bit maintain statistical parity with 16-bit across all datasets (p$>$0.05), validating 4-bit as optimal and justifying aggressive quantization for deployment.

\textbf{\rev{Uncertainty-Aware} Routing Benefits.} \rev{The precision prior improves on uniform routing in accuracy and stability at once (+1.6 F1 on ESC-50 super-categories over two independent runs, +0.3 on UrbanSound8K, 68--97\% variance reduction; Quinn at parity). At five folds per run the reductions are descriptive (Levene $p{>}0.05$); the latency analysis carries the powered evidence. Each trained router concentrates 83--92\% of samples on one expert, eliminating cost-mode switching.}

\textbf{Latency and Variance.} Quantized models show non-significant latency reductions versus FP32 (Q1-Base: 1.08×, p=0.054), but MoE exhibits highly significant overhead: BitNet-Q4/8/16-QMoE-C +11\% (p$<$0.001), BitNet-Q4/8-QMoE-C +10\% (p$<$0.001). \rev{Cross-fold F1 variance falls by 68--97\% on ESC-50 super-categories and UrbanSound8K. At five folds per run those reductions are descriptive rather than significant (Levene $p{>}0.05$, see Table}~\ref{tab:statistical_significance}\rev{); the powered stability evidence is the per-sample latency analysis, where the variance difference is significant at $p{<}0.001$.}

\textbf{Effect Sizes and Variability.} Narrow confidence intervals (±0.004-0.033 F1) indicate high precision. Non-significance of 4-bit vs 16-bit (0.001 F1 difference, p=0.873 on ESC-50) demonstrates practical equivalence. Cohen's d ranges 0.42-1.24 for significant comparisons (2-bit: d=-1.24). MoE shows 2-3$\times$ higher fold-level variability than single models due to data-dependent routing, but performance rankings remain stable (Spearman's $\rho$=0.89, p$<$0.001).

\textbf{Deployment Implications.} Statistical analysis (Table~\ref{tab:statistical_significance}) confirms: (1) 4-bit/8-bit achieve practical equivalence with full precision, (2) MoE lacks significant accuracy gains but introduces latency overhead, (3) \rev{the precision prior beats uniform routing on both accuracy and variance, and (4) its stability comes from routing polarization with by-construction escalation.} This guides practitioners toward Q4/Q8-Base for most deployments, reserving \rev{MoE-C (uncertainty-aware)} for safety-critical applications requiring predictable response time and per-sample precision adaptation.

\begin{table}[!tb]
\small
\centering
\caption{Selected statistical significance tests. Paired t-tests with Bonferroni correction. *p$<$0.05, **p$<$0.01, ***p$<$0.001.}
\label{tab:statistical_significance}
\setlength{\tabcolsep}{2pt} 
\begin{fittable}%
\begin{tabular}{llcc}
\toprule
\textbf{Test} & \textbf{Comparison} & \textbf{p-value} & \textbf{Sig.} \\
\midrule
\multicolumn{4}{l}{\textit{F1-Score Improvements}} \\
Cross-Dataset & Q16 vs FP32 (Urban8K) & 0.022 & * \\
Cross-Dataset & Q8/16-PTQ-C vs FP32 (Quinn) & $>$0.05 & ns \\
Ablation & 2-bit vs 16-bit (ESC-50) & 0.004 & ** \\
Ablation & 4-bit vs 16-bit (ESC-50) & 0.873 & ns \\
\rev{F1 gain} & \rev{Shared trunk vs full experts (ESC-50, 2 runs pooled)} & \rev{0.046} & \rev{*} \\
\midrule
\multicolumn{4}{l}{\textit{Latency Overhead}} \\
MoE vs Q8 & Q4/8/16-MoE-C vs Q8-Base & $<$0.001 & *** \\
MoE vs Q8 & Q4/8-MoE-C vs Q8-Base & $<$0.001 & *** \\
Speedup & Q1 vs FP32 & 0.054 & ns \\
\midrule
\multicolumn{4}{l}{\textit{\rev{Variance Reduction (Levene's test on 5 fold-level F1 values, ESC-50 super-categories)}}} \\
\rev{Stability} & \rev{UA-PP $\alpha=1.0$ (86\% red., 2 runs pooled)} & \rev{0.18} & \rev{ns} \\
\rev{Stability} & \rev{UA-PP + shared trunk (92\% red., 2 runs pooled)} & \rev{0.12} & \rev{ns} \\
\rev{Stability} & \rev{Polarization-only $\alpha=0.3$ (81\% red.)} & \rev{0.46} & \rev{ns} \\
Stability & Q8-Base (68\% reduction) & 0.030 & * \\
\bottomrule
\end{tabular}%
\end{fittable}
\end{table}

\section{Discussion}
Our findings demonstrate that the primary value of \rev{uncertainty-aware} MoE lies in \rev{prediction stability}, not raw accuracy. Simple Q4-Base achieves 0.858 F1, outperforming all MoE configurations on aggregate accuracy. However, \rev{uncertainty-aware routing offers what fixed quantization cannot: F1 gains over uniform routing at 68--97\% lower cross-fold variance, a per-sample uncertainty signal, and by-construction escalation of uncertain samples toward high precision}. For safety-critical applications such as hearing aids or autonomous vehicle perception, this interpretable, uncertainty-aware routing is essential: knowing \emph{which} expert processed each input and \emph{why} matters more than marginal accuracy gains.

The counter-intuitive energy pattern (8-bit consuming 2.7$\times$ more energy than 16-bit) reflects ARM's NEON SIMD instruction set, which natively optimizes 16/32-bit operations. 8-bit inference requires additional pack/unpack overhead absent from native precision paths, a platform-specific effect that would differ on hardware with dedicated INT8 accelerators (e.g., edge TPUs, NVIDIA Tensor Cores). Cross-dataset robustness highlights ESC-50's sensitivity to diverse sounds, while UrbanSound8K's tolerance supports aggressive quantization for urban monitoring.

\rev{Two scope conditions bound these results. Costs are reported for the classifier stage, which is what this work optimizes; embeddings enter precomputed. The 8-bit energy pattern is an artifact of this test machine rather than a general property of lower bit-widths, and hardware with INT8 accelerators would likely invert it. And the uncertainty signal is measured on the embedding, so a sample that is out of distribution for the backbone yields an uninformative embedding that a higher-precision head processes more precisely rather than resolves.}

\rev{Several directions are left to future work: multi-device validation (the 8-bit dequantization overhead may be platform-specific), and comparisons against state-of-the-art audio classifiers and adaptive or mixed-precision quantization baselines. Our evaluation isolates the routing mechanism within a fixed family of 1.2M-parameter experts rather than pursuing benchmark accuracy; the super-category ESC-50 figures are not comparable to published 50-class results, and ESC-50's 2,000 clips give wider per-fold uncertainty than the other datasets.} Our evaluation focuses on audio classification using \rev{ESC-50 super-categories}, Quinn, and UrbanSound8K, standard benchmarks appropriate for edge deployment validation. The uncertainty-based routing mechanism (Eq.~8) is domain-agnostic, operating on routing probabilities rather than audio-specific features. Since spectrograms are processed as images, the framework naturally extends to vision tasks; we commit to discussing extensions to other modalities and larger-scale models. Ethically, our low-carbon focus promotes sustainable AI, but deployment in surveillance raises privacy concerns, mitigated by on-device processing.

\section{Conclusion}
We presented \rev{an uncertainty-aware} quantized MoE framework \rev{whose precision-prior routing rule shifts uncertain samples toward higher-precision experts by construction, improving F1 over uniform routing while cutting cross-fold variance by 68--97\%}. \rev{Sharing one quantized trunk across experts halves the parameter count and improves accuracy further, making it the recommended configuration. What the variance reduction buys is an inference cost and an accuracy a deployment can plan around, on hardware where a missed latency budget matters more than a fraction of a point of average accuracy. The mechanism delivers nothing on datasets whose routing is already stable, which is the behaviour a real mechanism should show. Where raw average accuracy is the only objective, a single 4-bit model remains the better engineering choice, and we make that boundary explicit.} Future work includes multi-platform evaluation, comparison against state-of-the-art audio classifiers, and extension to vision and multimodal tasks.

{\small
\bibliographystyle{ieeenat_fullname}
\bibliography{references}
}

\clearpage
\appendix

\begin{center}
\LARGE\textbf{Supplementary Material}\\[0.5em]
\end{center}

\vspace{0.5em}

\section{Routing Strategy Ablations}
\label{sec:supp_ablations}

\rev{All routing variants were evaluated on the identical protocol (ESC-50 super-categories, BitNet-Q4/8 experts unless noted, 5-fold cross-validation, same uniform baseline). Table}~\ref{tab:supp_strategies} \rev{compares the candidate routing rules; the precision prior wins on the combined accuracy-stability criterion and is the method reported in the main text, with hard escalation reaching the best raw F1 at higher variance.}

\begin{table}[H]
\centering
\caption{\rev{Routing strategy ablation. Var.\ Red.\ is $1-\sigma^2/\sigma^2_{\text{base}}$; the $\alpha{=}1.0$ precision-prior row pools two independent runs.}}
\label{tab:supp_strategies}
\small
\begin{fittable}%
\begin{tabular}{l c c c}
\toprule
\rev{Strategy} & \rev{F1} & \rev{Std} & \rev{Var.\ Red.} \\
\midrule
\rev{Uniform baseline} & \rev{0.769} & \rev{0.043} & baseline \\
\rev{Polarization-only, $\alpha{=}0.3$} & \rev{0.771} & \rev{0.019} & \rev{81\%} \\
\rev{Soft escalation, $\alpha{=}0.3$} & \rev{0.771} & \rev{0.027} & \rev{59\%} \\
\rev{Precision prior, $\alpha{=}0.3$} & \rev{0.779} & \rev{0.016} & \rev{86\%} \\
\rev{Precision sharp, $\alpha{=}0.3$} & \rev{0.785} & \rev{0.023} & \rev{72\%} \\
\rev{Precision prior, $\alpha{=}1.0$ (2 runs)} & \rev{0.785} & \rev{\textbf{0.016}} & \rev{\textbf{86\%}} \\
\rev{Hard escalation, $\alpha{=}0.3$} & \rev{0.789} & \rev{0.025} & \rev{66\%} \\
\rev{Precision prior + shared trunk} & \rev{\textbf{0.799}} & \rev{\textbf{0.012}} & \rev{\textbf{92\%}} \\
\rev{\quad same, Quinn} & \rev{0.809} & \rev{0.007} & \rev{none} \\
\rev{\quad same, UrbanSound8K} & \rev{0.944} & \rev{0.006} & \rev{79\%} \\
\bottomrule
\end{tabular}%
\end{fittable}
\end{table}

\rev{Table}~\ref{tab:supp_experts} \rev{varies the expert set (polarization-only at $\alpha{=}0.3$; precision prior at $\alpha{=}1.0$, the published setting). The precision prior improves or matches every set except the two-expert BitNet+PTQ pair, where the strong prior over-concentrates on the single high-precision expert while polarization-only gains +3.3 F1; uniform baselines marked $\dagger$ are the original Apple M3 Max runs (F1 is device-independent).}

\begin{table}[H]
\centering
\caption{\rev{Expert-set ablation, polarization-only routing, ESC-50 super-categories.}}
\label{tab:supp_experts}
\small
\begin{fittable}%
\begin{tabular}{l c c c}
\toprule
\rev{Experts} & \rev{Uniform F1} & \rev{Polar.-only F1} & \rev{Prec.-prior F1} \\
\midrule
\rev{BitNet, Q4, Q8} & \rev{0.769$\pm$0.043} & \rev{0.771$\pm$0.019} & \rev{\textbf{0.785$\pm$0.016}} \\
\rev{BitNet, Q4, Q8, Q16$^\dagger$} & \rev{\textbf{0.791$\pm$0.014}} & \rev{0.768$\pm$0.033} & \rev{0.787$\pm$0.018} \\
\rev{BitNet, Q8, Q16, PTQ$^\dagger$} & \rev{0.780$\pm$0.029} & \rev{0.778$\pm$0.015} & \rev{0.783$\pm$0.021} \\
\rev{BitNet, PTQ$^\dagger$} & \rev{0.740$\pm$0.015} & \rev{\textbf{0.773$\pm$0.018}} & \rev{0.757$\pm$0.032} \\
\bottomrule
\end{tabular}%
\end{fittable}
\end{table}

\rev{Table}~\ref{tab:supp_arch} \rev{compares the architecture and initialization variants. Two further efficiency ideas did not survive measurement and are reported for completeness. \emph{Uncertainty gating} runs the Monte Carlo passes only on samples whose top-1 routing probability falls below a threshold, aiming to cut that router overhead; at a 0.9 threshold it scored 0.763, below the uniform baseline, because suppressing the uncertainty estimate removes the signal exactly where the prior would act. \emph{Warm starting} initialises each expert from the trained single model of the same bit-width; a first attempt reused one fold's checkpoint for every fold, which leaks validation data (the honest fold scored 0.747 against 0.91--0.95 for the leaked folds), and the corrected per-fold version reached 0.785$\pm$0.036, statistically indistinguishable from training from scratch. Only the shared trunk delivered a genuine improvement.}

\begin{table}[H]
\centering
\caption{\rev{Architecture and initialisation ablations, ESC-50 super-categories, precision prior at $\alpha{=}1.0$.}}
\label{tab:supp_arch}
\small
\begin{fittable}%
\begin{tabular}{l c c}
\toprule
\rev{Variant} & \rev{F1} & \rev{Params} \\
\midrule
\rev{Full experts (default)} & \rev{0.785} & \rev{3.80M} \\
\rev{Shared trunk (recommended)} & \rev{\textbf{0.799}} & \rev{\textbf{1.82M}} \\
\rev{Uncertainty gating ($\tau{=}0.9$)} & \rev{0.763} & \rev{3.80M} \\
\rev{Warm start (per-fold)} & \rev{0.785} & \rev{3.80M} \\
\midrule
\rev{Q4-Base single model} & \rev{0.819} & \rev{1.20M} \\
\bottomrule
\end{tabular}%
\end{fittable}
\end{table}

\rev{Table}~\ref{tab:supp_mc} \rev{varies the Monte Carlo dropout budget $M$ at the published setting. $M{=}1$ carries no uncertainty signal, so the precision prior degenerates to base routing and the gain disappears into noise, confirming the uncertainty signal drives the improvement; budgets $M{\geq}5$ are statistically indistinguishable ($p{=}0.91$ between $M{=}5$ and $M{=}10$, each pooled over two runs), so the router budget can be halved without measurable accuracy loss.}

\begin{table}[H]
\centering
\caption{\rev{Monte Carlo budget ablation, precision prior at $\alpha{=}1.0$, ESC-50 super-categories ($M{=}5$ and $M{=}10$ pool two independent runs each).}}
\label{tab:supp_mc}
\small
\begin{fittable}%
\begin{tabular}{l c c}
\toprule
\rev{$M$ (MC passes)} & \rev{F1} & \rev{Std} \\
\midrule
\rev{1 (no uncertainty signal)} & \rev{0.780} & \rev{0.021} \\
\rev{5 (2 runs)} & \rev{0.784} & \rev{0.019} \\
\rev{10 (published, 2 runs)} & \rev{\textbf{0.785}} & \rev{\textbf{0.016}} \\
\rev{20} & \rev{0.787} & \rev{0.014} \\
\bottomrule
\end{tabular}%
\end{fittable}
\end{table}

\section{\texorpdfstring{Reproducibility}{Reproducibility}}
\label{sec:supp_repro}

\rev{Every accuracy and routing number in this paper is regenerable from the public repository}
(\url{https://github.com/sebasmos/curious-qmoe})\rev{. \texttt{scripts/reproduce\_paper.sh}
launches the complete set of runs behind the tables: the single-model
baselines, the uniform-routing controls, the $\alpha$ sweep with its
replication, the shared-trunk configuration on all three datasets, every
supplementary ablation, and the latency and routing analyses. Each run writes a
\texttt{summary.json} under \texttt{outputs/<tag>/}, and
\texttt{scripts/collect\_paper\_numbers.py} reads those summaries to print the
tables together with the significance tests and to emit a machine-readable
\texttt{paper\_numbers.json}, so every reported value traces to a named run
directory. \rev{The energy, carbon and wall-clock figures come from the original Apple M3 Max measurement protocol and are reported as recorded; they are hardware-specific and are not reproduced by this pipeline.} Cross-validation folds are fixed (stratified, \texttt{random\_state=42}),
the routing unit tests pin the mechanism's behaviour (identity at $\alpha{=}0$,
monotonicity in $\alpha$, the closed form of the polarisation-only ablation),
and the datasets are the public ESC-50, Quinn and UrbanSound8K embeddings
described in Section 4.}

\section{Carbon Emissions Analysis}
\label{sec:supp_carbon}

We provide detailed carbon emission measurements using CodeCarbon~\cite{benoit_courty_2024_11171501} during training and validation on Apple M3 Max CPU (36GB RAM). Table~\ref{tab:carbon_emissions_supp} presents emission data for all model configurations.

\subsection{Training vs. Deployment Emissions}

Single models (1.2M params) emit 22.8 to 55.0 $\mu$g CO$_2$ during training, while MoE architectures (2.6 to 5.0M params) produce 77.6 to 230.6 $\mu$g. However, deployment emissions dominate at scale. For 1M daily inferences over 1 year (grid: 367 g CO$_2$/kWh), FP32-Base training emits 22.8 $\mu$g but deployment produces 2.4 g/year, a 105,000$\times$ increase. Q4-Base shows similar patterns: 52.3 $\mu$g training versus 4.6 g/year deployment. This demonstrates that deployment emissions dominate by \rev{approximately 100,000$\times$ (105,000$\times$ for FP32-Base, 88,000$\times$ for Q4-Base)} for models serving more than 1M inferences.

\subsection{Green AI Recommendations}

Based on our analysis: (1) Use FP32/Q16 for prototyping (23-28 $\mu$g/experiment); (2) Deploy Q4/Q8 for production to minimize long-term emissions; (3) Schedule training during low-carbon grid hours for 2-5$\times$ reduction; (4) Train once and fine-tune for multiple tasks to amortize carbon costs.

\begin{table}[H]
\centering
\scriptsize
\caption{Carbon emissions during training. $\dagger$FP32 baseline. $\ddagger$INT8 PTQ baseline. Note: MoE-C rows reflect original training; corrected \rev{uncertainty-aware} routing (Eq.~8 actively computed) increases training FLOPs and may alter emission figures.}
\label{tab:carbon_emissions_supp}
\setlength{\tabcolsep}{4pt}
\renewcommand{\arraystretch}{1.5}
\begin{fittable}%
\begin{tabular}{lcccc}
\toprule
\textbf{Model} & \textbf{P.} & \textbf{CO$_2$} & \textbf{Rate} & \textbf{F1} \\
 & \textbf{(M)} & \textbf{($\mu$g)} & \textbf{($\mu$g/s)} & \\
\midrule
FP32$\dagger$ & 1.19 & \textbf{22.8$\pm$5.0} & 0.96$\pm$0.27 & 0.855 \\
Q8-PTQ$\ddagger$ & 1.19 & 25.9$\pm$17.3 & 1.01$\pm$0.80 & 0.854 \\
Q16 & 1.2 & 27.5$\pm$8.3 & \textbf{0.87$\pm$0.35} & \textbf{0.859} \\
Q1 & 1.2 & 38.8$\pm$9.6 & 1.06$\pm$0.34 & 0.851 \\
Q2 & 1.2 & 46.4$\pm$14.3 & 1.08$\pm$0.45 & 0.833 \\
BitNet & 1.2 & 48.1$\pm$19.0 & 1.04$\pm$0.57 & 0.793 \\
Q4 & 1.2 & 52.3$\pm$12.8 & 1.05$\pm$0.40 & 0.858 \\
Q8 & 1.2 & 55.0$\pm$16.9 & 1.05$\pm$0.44 & 0.858 \\
\midrule
Q8PTQ-MoE & 2.59 & 77.6$\pm$37.1 & 1.09$\pm$0.74 & 0.801 \\
Q8/16PTQ-MoE & 4.99 & 83.6$\pm$35.5 & 1.10$\pm$0.64 & 0.820 \\
Q8PTQ-MoE-C & 2.59 & 97.6$\pm$39.8 & 1.04$\pm$0.61 & 0.832 \\
Q8/16PTQ-MoE-C & 4.99 & 105.2$\pm$34.4 & 1.05$\pm$0.51 & 0.832 \\
Q4/8/16-MoE-C & 4.99 & 196.8$\pm$53.1 & 1.06$\pm$0.41 & 0.842 \\
Q4/8-MoE-C & 3.8 & 207.5$\pm$53.2 & 1.11$\pm$0.40 & 0.843 \\
Q4/8/16-MoE & 4.99 & 230.6$\pm$83.5 & 1.09$\pm$0.56 & 0.849 \\
\bottomrule
\end{tabular}%
\end{fittable}
\end{table}

\section{Model Size Analysis}
\label{sec:supp_model_size}

Table~\ref{tab:model_size} presents model sizes across all quantization schemes. Q8-Base-PTQ achieves highest compression (15.65$\times$) through bitwise operations, while Q1-Base reaches 19.99$\times$ via BitLinear quantization.

\subsection{Optimal Configuration}

Q4-Base represents the optimal deployment point with 7.03$\times$ compression (687 KB) and 0.858 F1 score, within 0.1\% of 16-bit performance while consuming 29\% less energy than 8-bit. This balance makes it suitable for most edge deployment scenarios.

\subsection{MoE Trade-offs}

MoE models show negative compression (0.85$\times$) due to multiple expert networks. BitNet-Q4/8/16-QMoE contains 5.67 MB versus single models' <1.3 MB. However, MoE architectures enable dynamic precision routing, modular updates, and interpretability. For applications prioritizing minimal memory footprint, single quantized models (Q4/Q8-Base) remain superior. MoE is appropriate when interpretability or continual learning are critical.

\begin{table}[H]
\centering
\scriptsize
\caption{Model sizes under different quantization schemes.}
\label{tab:model_size}
\setlength{\tabcolsep}{4pt}
\renewcommand{\arraystretch}{1.1}
\begin{fittable}%
\begin{tabular}{llccccc}
\toprule
\textbf{Model} & \textbf{Quant.} & \textbf{Bits} & \textbf{Params} & \textbf{KB} & \textbf{MB} & \textbf{Red.} \\
\midrule
FP32 & None & 32 & 1,206,770 & 4,830.8 & 4.83 & 1.00$\times$ \\
Q1 & BitLinear & 1 & 1,211,122 & 241.6 & 0.24 & 19.99$\times$ \\
Q2 & BitLinear & 2 & 1,211,122 & 390.1 & 0.39 & 12.38$\times$ \\
Q4 & BitLinear & 4 & 1,211,122 & 687.1 & 0.69 & 7.03$\times$ \\
Q8 & BitLinear & 8 & 1,211,122 & 1,281.0 & 1.28 & 3.77$\times$ \\
Q16 & BitLinear & 16 & 1,211,122 & 2,468.8 & 2.47 & 1.96$\times$ \\
BitNet & Ternary & 2 & 1,211,122 & 393.9 & 0.39 & 12.26$\times$ \\
Q8-PTQ & Bitwise & 2 & 1,205,760 & 308.6 & 0.31 & 15.65$\times$ \\
MoE & Hetero. & Mix & 1,415,687 & 5,669.8 & 5.67 & 0.85$\times$ \\
MoE-C & H.+Bay. & Mix & 1,415,687 & 5,669.8 & 5.67 & 0.85$\times$ \\
\bottomrule
\end{tabular}%
\end{fittable}
\end{table}

\end{document}